# Truck-and-Trailer Backer-Upper problem using Cascaded Fuzzy Controllers

Yurui Ming

CIBCI Lab, School of Computer Science, FEIT, University of Technology Sydney, Australia

**Abstract.** In this paper we craft a cascaded fuzzy controlling system for the traditional Truck-and-Trailer Backer-Upper problem, which is a benchmarking for testing various intelligent controlling systems. Inspired by the most inclination of human operations, we decompose the original overall controlling problem into two sub-controlling problems. A first fuzzy controller which predicts the optimal deviation of the trailer leading to a most potential successful docking is designed based on previous work of other scholars, then a second fuzzy controller which maximizes such a potentiality is applied to the truck in an intuitive manner. The final simulation and results not only demonstrate the practicability of the approach proposed in this paper, but also exhibits the dramatic simplicity compared with previous work which try to optimize the system from overall perspective.

## 1   Introduction

The precise docking of a truck and trailer at a loading dock is a non-trivial task even for experienced drivers. Usually erroneous movements have been spotted like repeatedly going forward and backing again before the final docking. From its initial publishing, which was utilizing neural network [1], other approaches such as genetic programming [2] and fuzzy control [3] were proposed against this problem from time to time.

Fuzzy control has proven its feasibility to solve highly nonlinear systems exempted from complex mathematical modelling, which is exemplified by the Truck-and-Trailer Backer-Upper problem that has drawn lots of attention along the research. Besides the work mentioned above, other papers with insights from different aspects of fuzzy theory also investigated the problem [4-6]. However, due to the complexity of the truck-and-trailer system itself, the approaches proposed in the above papers tend to be too complicated which can easily burry the merit of fuzzy control theory, typically due to the so-called curse of dimensionality.

Actually, careful observations from practice can help reflecting the truck-and-trailer problem in some intuitive way. It can be spotted during the backing procedure, to directly back the whole system is of least preferred operation. Usually some pivot tends to be referred to firstly to adjust the trailer in an appropriate position, then operations continue applying to the steering wheel of the cab which maximize this kind of optimal or potential for a successful docking, even if the pivot may only exist virtually in mind based on estimation. From such an argument, in this paper we design a system in a decomposed way with each part corresponding to different stages mentioned above. Since each time we deal with a subsystem with less control variables, the general dimension challenge is avoided for fuzzy controller design when the effectiveness is still maintained.

## 2 System Design

The system is briefly depicted before detailing the procedures, especially about the symbols used in this work. Referring to Fig. 1 for illustration of the plant and Table 1 for explanation of the symbols. As mentioned above, in light of human's operation, the overall system will be divided into two cascaded parts to consider each objective of the sub-systems.

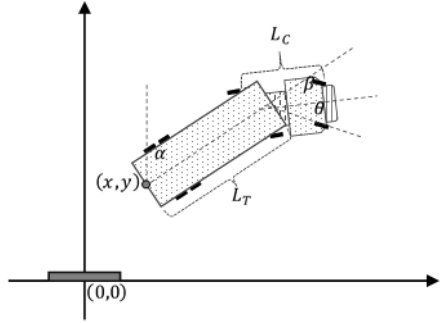

**Fig. 1.** Illustration of the of the Truck-and-Trailer Backer-Upper problem with corresponding symbols.

When isolating the trailer for examination, the aim is to obtain a $\beta'$ which can lead to the most desirable $\alpha$ updated in next iteration. Since now we consider the trailer is self-driven, but not by the force from the cab, or it is equivalent to treating the trailer as a "truck" or "car". Such a treatment can also harness the well-established results in other papers, because how to back a car has been studied thoroughly. Hence the designing of the first part is effective but effortless, and a fuzzy controller $FLC_T$ is devised for this purpose.

**Table 1.** Plant Description

| | | |
|---|---|---|
| State Variables | $(x, y)$ | Coordinates of the center rear of the trailer |
| | $\alpha$ | The angle between the $y$-axis and the deviation of the trailer |
| | $\beta$ | The angle between the deviation of the cab with regard to the direction of the trailer |
| Control Variable | $\theta$ | The angle between the deviation of the steering wheel with regard to the cab |
| Constrains | | $100 \leq x \leq 100, y \geq 0$ |
| | | $-180° \leq \alpha \leq 180°$ |
| | | $|\alpha - \beta| \leq 90°$ |
| | | $-30° \leq \theta \leq 30°$ |
| Parameters | $v$ | The velocity of the steering wheel during backing |
| | $L_C$ | Length of the cab |
| | $L_T$ | Length of the trailer |
| Objectives | | $(x, y) = 0, \alpha = 0$ |

It just emphasizes here that if only with a truck or car, $\beta$ can be changed flexibly at consecutive time samples. However, in the truck-and-trailer system, the range of cab's movement constrains the magnitude of how $\beta$ can change in each step.

As in practice, the second part of the system is to tailor $\theta$ for satisfying the $\beta'$ requested by the first controller. Just as mentioned, the consequent $\beta$ produced by adjusting $\theta$ may not coincident with the optimal variation $\beta'$, and an intuitive solution is to minimize the difference, indicating by the following model:

$$\min_{|\theta|\leq 30°}|\beta' - \beta| \tag{1}$$

The goal can be treated as the potentiality indicating a successful docking and a fuzzy controller $FLC_C$ will be designed to fulfil the aim.

## 2.1 Design of $FLC_T$

As mentioned, the design of $FLC_T$ is based on the previous work, such as in [3]. The details are omitted here for conciseness, just the brief description of member functions for input and output variables and the rule table.

For $x$ from displacement perspective, a linguistic distance variable $X$ is defined. The term set $T(X)$ contains the exact terms $\{LE, LC, CE, RC, RI\}$ with membership functions as in Fig. 3.

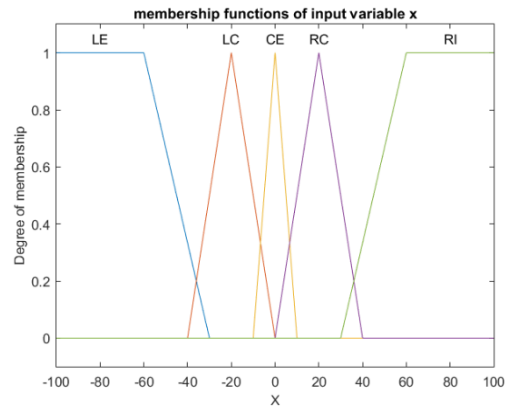

**Fig. 3.** Illustration of the designed membership functions of $x$.

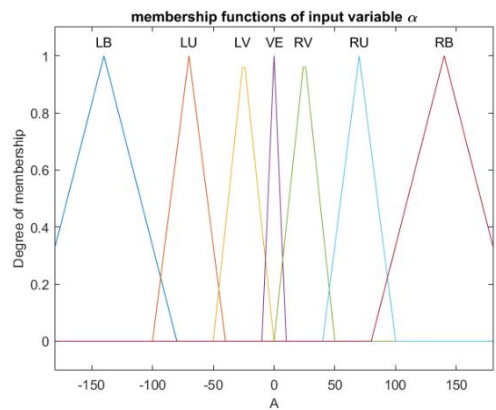

**Fig. 4.** Illustration of the designed membership functions of $\alpha$.

**Table 2.** Rule enumeration for $FLC_T$

|    | LE | LC | CE | RC | RI |
|----|----|----|----|----|----|
| **LB** | PS | PM | NS | NM | NB |
| **LU** | NS | PS | PM | PB | PB |
| **LV** | NS | NS | PS | PM | PB |
| **VE** | NM | NS | ZE | PS | PM |
| **RV** | NB | NM | NS | PS | PS |
| **RU** | NB | NB | NM | NS | PS |
| **RB** | PB | PM | PS | NM | NS |

For $y$ it is out of consideration for controlling in this work so there's no bother for defining a corresponding linguistic variable.

For $\alpha$ from deviation perspective, a linguistic direction variable $A$ is defined. The term set of $A$, namely $T(A)$ contains the exact terms $\{LB, LU, LV, VE, RV, RU, RB\}$ with the membership functions as in Fig. 4.

As aforementioned, the variable $\beta$ is the output of the $FLC_T$. This paper is based on the previous work of other scholars to take the advantage of well-defined rules, which are keys to the performance of $FLC_T$. It does not indicate the limitation of the method covered in this paper, but just saving the effort in consulting experts for constructing better rules. Here the range of $\beta$ is restricted from general $[-90°, 90°]$ to $[-30°, 30°]$ relative to the direction of the trailer, for better coordinating with the controller $FLC_C$ mentioned below.

So, a linguistic variable $B$ is defined with the term set $T(B)$ contains the usual polarity term $\{NB, NM, NS, ZE, PS, PM, PB\}$ and the corresponding membership functions in Fig. 5.

Finally, rules for $FLC_T$ is self-described as in Table 2.

## 2.2 Design of $FLC_C$

As mentioned above, the goal of the second controller is to solve (1), aka minimize the difference between the actual $\beta$ with the expectation $\beta'$, in order to maintain the potentiality for a successful docking. A variable $\gamma \equiv \beta' - \beta$ is defined to aid the analysis. Since $\beta$ has been restricted to $[-30°, 30°]$ relative to the direction of the trailer, so $\gamma$ belongs to $[-60°, 60°]$.

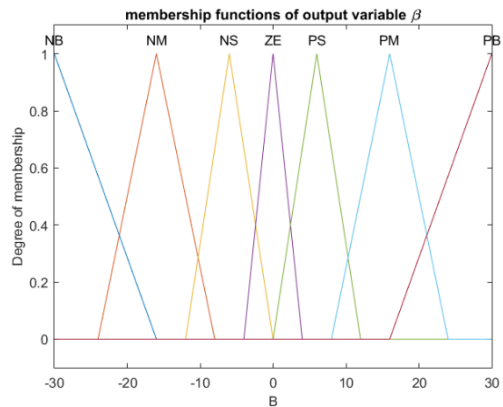

**Fig. 5.** Illustration of the designed membership functions of $\beta$.

A linguistic polarity variable $G$ is defined for $\gamma$ with the term set $T(G)$ contains the usual polarity term $\{NB, NM, NS, ZE, PS, PM, PB\}$ as above. The membership functions are shown in Fig. 6.

For $FLC_C$ the output is $\theta$ to adjust steering, based on which the whole system will be updated, and new values will be feedbacked for new iterations. A linguistic steering variable $S$ is defined with $T(S)$ contains terms as $\{NB, NM, NS, ZE, PS, PM, PB\}$. The membership functions are as in Fig. 7.

The rule for $FLC_C$ is clearly intuitive and simple, just take the best strategy to minimize $\gamma$. There's no surprise it's just consisted by a simple table even seems trivial as in Table 3.

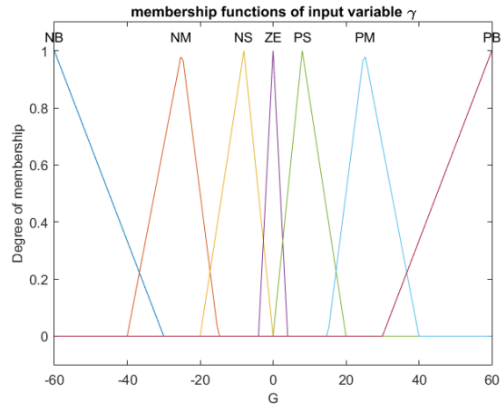

**Fig. 6.** Illustration of the designed membership functions of $\gamma$.

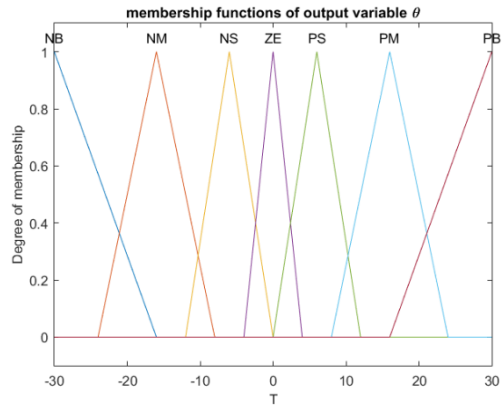

**Fig. 7.** Illustration of the designed membership functions of $\theta$.

**Table 3.** Rule enumeration for $FLC_C$

| NB | NM | NS | ZE | PS | PM | PB |
|----|----|----|----|----|----|----|
| NB | NM | NS | ZW | PS | PM | PB |

## 3 Simulation

Although an exact mathematical modelling is exempted when applying fuzzy controlling, however since for simulation, the state variables of each step should be derived from the system in order to drive the consequent following-up, so approximating dynamic equations are still required for correlating the relation between input and output. Based on description in above sections, the total system is modeled by taking advantage of Simulink as in Fig. 8.

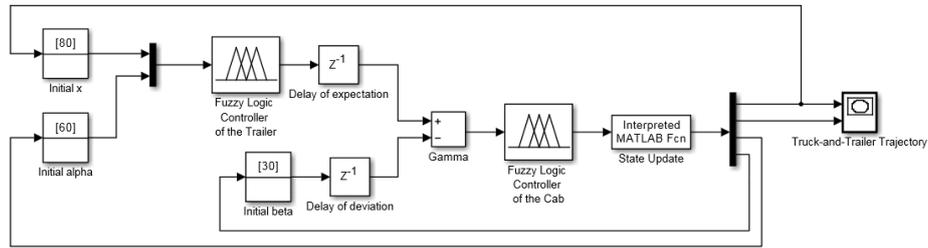

**Fig. 8.** System design via Simulink

In this work, for simplicity, the following equations are utilized for simulation:

$$d_c[t] = -v * \cos(\theta[t]) \tag{2}$$

$$d_t[t] = d_c[t] * \cos(\beta[t]) \tag{3}$$

$$x[t+1] = x[t] + d_t[t] * \sin\alpha[t] \tag{4}$$

$$y[t+1] = y[t] + d_t[t] * \cos\alpha[t] \tag{5}$$

$$\alpha[t+1] = \alpha[t] - \arcsin^{-1}\left(\frac{d_c * \sin(\beta[t])}{L_t}\right) \tag{6}$$

$$\beta[t+1] = \beta[t] - \arcsin^{-1}\left(\frac{-v * \sin(\theta[t])}{L_c}\right) \tag{7}$$

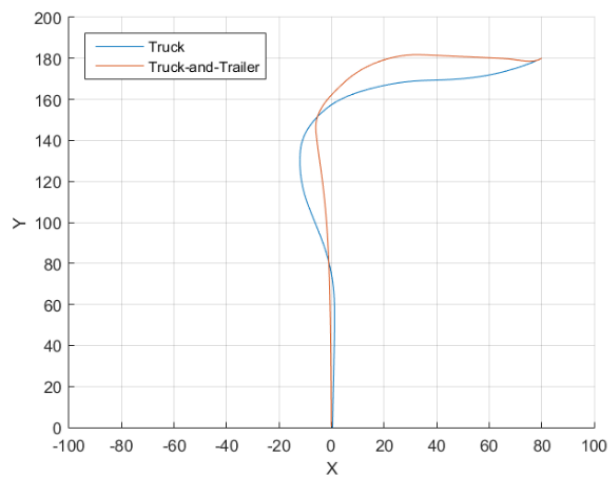

**Fig. 9.** $v = 1$, $L_C = 2$, $L_T = 8$, $(x, y) = (80, 180)$, $\alpha = 60°$, $\beta = 30°$.

The meanings of the variables have explained in TABLE I. $d_c$ and $d_t$ are distances of the movement of the cab and trailer at each time step, respectively. It is also supposed during consecutive time sampling; the velocity of steering wheel is the constant $v$.

Various cases have been verified, including the cases when there is insufficient space between x-axis which lead a docking failure. It is observed for most cases, as long as the $FLC_T$ is with appropriate initial state, the system will succeed. Several cases are enlisted for demonstrating the results. During each case, the trajectories of solely running $FLC_T$ and running the whole system respectively are plotted in the same figure for comparison, as illustrated in Fig. 9 to Fig. 11.

In fact, the performance of the approach in this work lays on two factors. One is $FLC_T$ which is rely on previous work of other scholars; the other is the $FLC_c$ which is based on the intuitive treatment proposed in this paper and is the key measurement here. In all cases above, since two trajectories tend to converge with time evolving, which is a great indicator showing the efficiency and simplicity of our method.

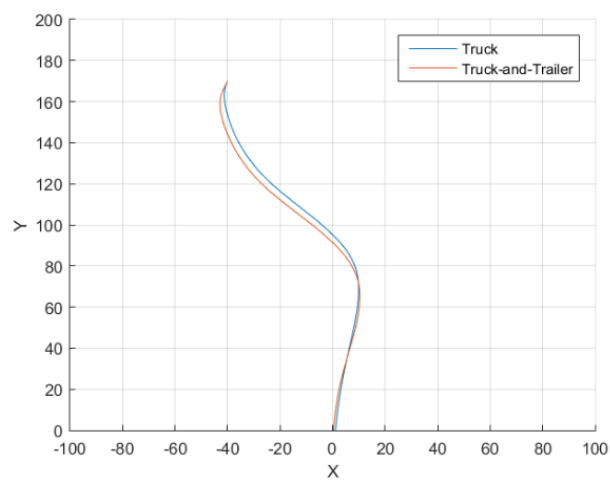

**Fig. 10.** $v = 1$, $L_C = 2$, $L_T = 8$, $(x, y) = (-40, 170)$, $\alpha = 20°$, $\beta = 15°$.

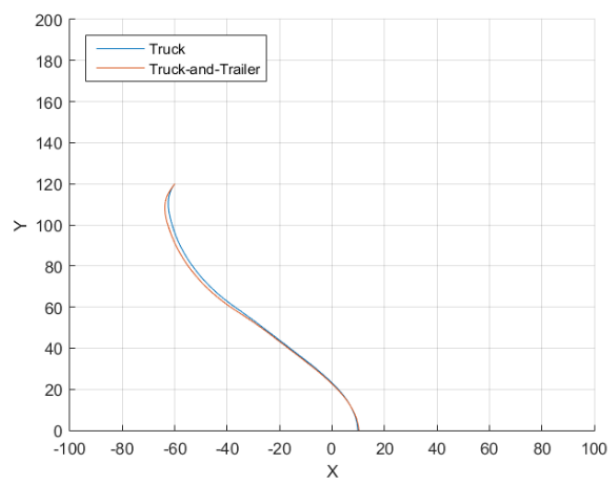

**Fig. 11.** $v = 1$, $L_C = 2$, $L_T = 8$, $(x, y) = (-60, 120)$, $\alpha = 30°$, $\beta = 0°$.

## 4      Conclusion

In this paper, an approach utilizing cascaded fuzzy controllers is proposed for solving the traditional Truck-and-Trailer Backer-Upper problem. By taking advantage of the previous work and intuition from practice, two controllers are devised for solving objectives in a joint way. The results of the simulation show the practicability of the method, and its simplicity is reflected during the design and implementation. It demonstrates an effective but simple design of fuzzy control system can still maintain high performance. Such an approach is worthwhile being transferred and generated to solve similar problems with high non-linear properties.